\documentclass[letterpaper]{article} 
\usepackage{aaai24}  
\usepackage{times}  
\usepackage{helvet}  
\usepackage{courier}  
\usepackage[hyphens]{url}  
\usepackage{graphicx} 
\usepackage{natbib}  
\usepackage{caption} 
\usepackage{algorithm}
\usepackage{algorithmic}
\usepackage{multirow}
\usepackage{booktabs} 
\usepackage{newfloat}
\usepackage{listings}
\DeclareCaptionStyle{ruled}{labelfont=normalfont,labelsep=colon,strut=off} 
\floatstyle{ruled}
\newfloat{listing}{tb}{lst}{}
\floatname{listing}{Listing}
\title{Defying Imbalanced Forgetting in Class Incremental Learning}
\author {
    Shixiong Xu\textsuperscript{\rm 1,2},
    Gaofeng Meng\textsuperscript{\rm 1,2,3}\thanks{Corresponding author.},
    Xing Nie\textsuperscript{\rm 1,2},
    Bolin Ni\textsuperscript{\rm 1,2},
    Bin Fan\textsuperscript{\rm 4},
    Shiming Xiang\textsuperscript{\rm 1,2}
}
\affiliations {
    \textsuperscript{\rm 1}State Key Laboratory of Multimodal Artificial Intelligence Systems, Institute of Automation, Chinese Academy of Sciences\\
    \textsuperscript{\rm 2}School of Artificial Intelligence, University of Chinese Academy of Sciences\\
    \textsuperscript{\rm 3}Centre for Artificial Intelligence and Robotics, HK Institute of Science \& Innovation, Chinese Academy of Sciences\\
    \textsuperscript{\rm 4}School of Intelligence Science and Technology, University of Science and Technology, Beijing
\\
    
    \{xushixiong2020, niexing2019, nibolin2019\}@ia.ac.cn, 
    bin.fan@ieee.org,
    \{gfmeng, smxiang\}@nlpr.ia.ac.cn
}

\begin{document}

\maketitle

\begin{abstract}
We observe a high level of imbalance in the accuracy of different classes in the same old task for the first time. This intriguing phenomenon, discovered in replay-based Class Incremental Learning (CIL), highlights the imbalanced forgetting of learned classes, as their accuracy is similar before the occurrence of catastrophic forgetting. This discovery remains previously unidentified due to the reliance on average incremental accuracy as the measurement for CIL, which assumes that the accuracy of classes within the same task is similar. However, this assumption is invalid in the face of catastrophic forgetting. Further empirical studies indicate that this imbalanced forgetting is caused by conflicts in representation between semantically similar old and new classes. These conflicts are rooted in the data imbalance present in replay-based CIL methods. Building on these insights, we propose CLass-Aware Disentanglement (CLAD) to predict the old classes that are more likely to be forgotten and enhance their accuracy. Importantly, CLAD can be seamlessly integrated into existing CIL methods. Extensive experiments demonstrate that CLAD consistently improves current replay-based methods, resulting in performance gains of up to 2.56\%.
\end{abstract}

\section{Introduction}
\label{sec:intro}
In typical image recognition tasks, the data is assumed to follow the independently and identically distributed (i.i.d.) assumption. A good classification model is expected to have similar and high accuracy across different classes. But in the real world, the data is non-stationary. To address this issue, Class Incremental Learning allows the model to continually learn new classes without forgetting the previously learned ones~\cite{three_scenarios}. However, when sequential fine-tuning is performed on new classes without the presence of old data, there is a dramatic drop in accuracy for the learned tasks, known as catastrophic forgetting~\cite{catastrophic_forgetting}.

\begin{figure}[ht]
  \centering
   \includegraphics[width=1.0\linewidth]{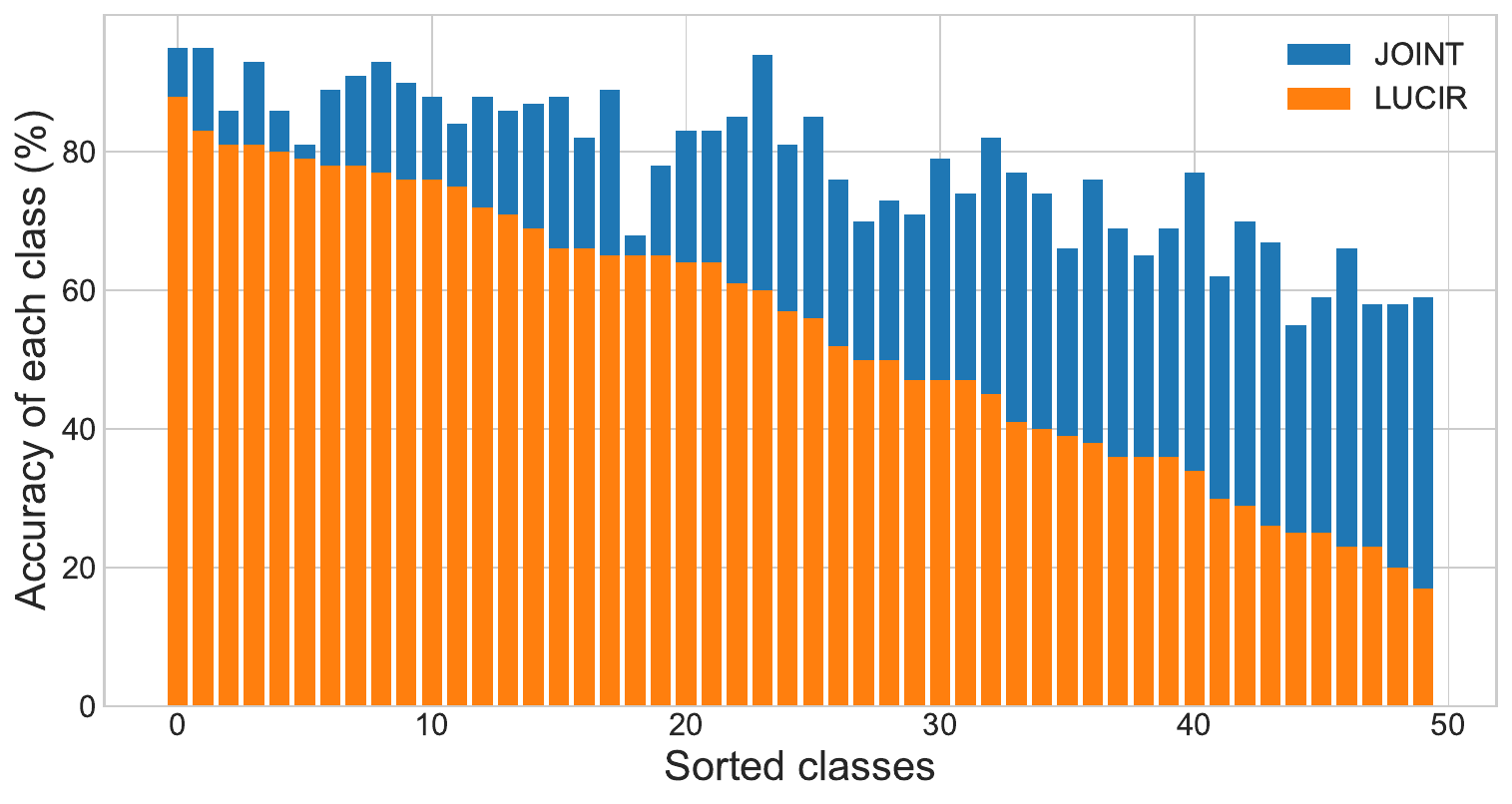}

   \caption{Demonstration of imbalanced forgetting. Visualization of the accuracy of each class in the first task obtained by joint training and LUCIR~\cite{LUCIR}. The class indexes are sorted according to the result from LUCIR for better visualization.More illustrations of the above phenomenon with other methods can be found in the supple- mentary material.}
   \label{fig:imbalanceforgetting}
\end{figure}

As the accuracy between the old and new classes is highly imbalanced in CIL, numerous approaches have been proposed. Among them, exemplar replay has been proven to be a simple yet effective strategy ~\cite{ER,ICARL,MER,DER}. In exemplar replay, a subset of each class is selected and stored in a buffer. During the training of subsequent tasks, these exemplars are reused in various ways to help preserve the learned knowledge, such as joint training~\cite{ER,MER}, knowledge distillation~\cite{LUCIR,ICARL}, gradient projection~\cite{GEM,GPM,FSDGPM}, and bias correction~\cite{LUCIR, BiC, WA, GDumb}. It is important to note that all these efforts primarily focus on tackling the accuracy imbalance between old and new classes. However, we argue that these efforts alone are insufficient to meet the expectations of a classification model.

\begin{figure*}[t]
  \centering
   \includegraphics[width=1.0\linewidth]{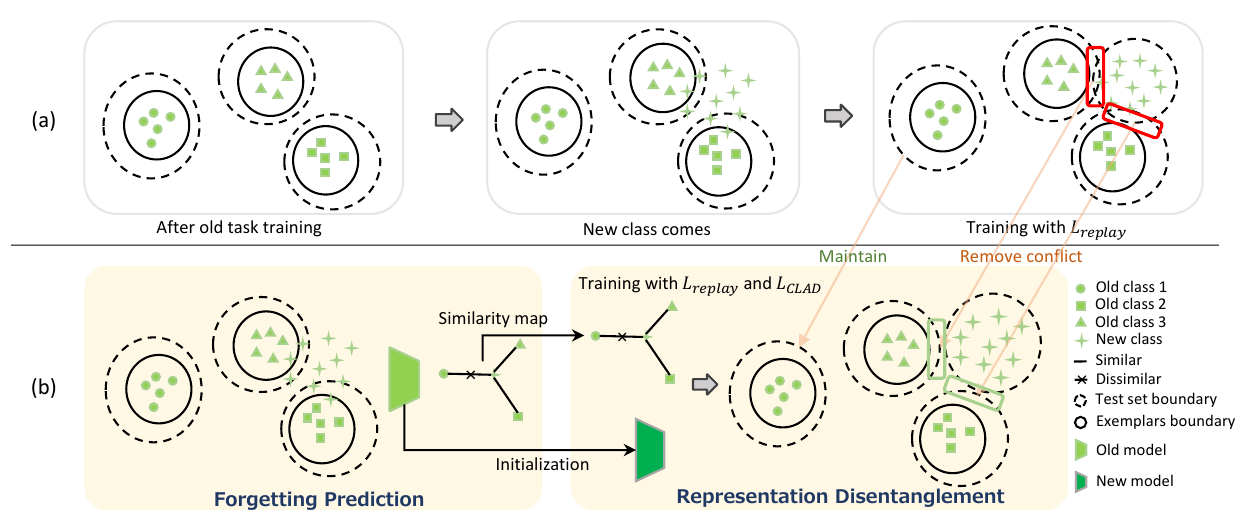}

   \caption{An overview of (a) existing replay-based methods and (b) our proposed CLAD. In existing replay-based methods, different old classes 1,2, and 3 have different accuracy because of the different similarities with the new class. The limited exemplars are not sufficient to preserve the boundary of the test set (low accuracy of classes 2 and 3 in (a)). Our proposed CLAD consists of two parts: Forgetting Prediction (FP) and Representation Disentanglement (RD). FP aims to find the classes that might be forgotten during the learning of new classes (classes 2 and 3). Based on the similarity information from FP, RD encourages the representation of new classes to stay away from similar old ones.}
   \label{fig:overview}
\end{figure*}

For the first time, we observed that the accuracy between classes of the same old task is also highly imbalanced.  It is evident that \textbf{imbalanced forgetting} occurs during the learning of a new task, as their accuracy is similar just after learning (as shown in Fig.~\ref{fig:imbalanceforgetting} JOINT). This phenomenon remains undiscovered because the average incremental accuracy used for measuring CIL approaches assumes that the accuracy of the classes within the same task is uniform. Fig.~\ref{fig:imbalanceforgetting} provides an example where the model is trained using a CIL setting that splits CIFAR-100~\cite{CIFAR100} into six tasks, with 50 classes in the first task and 10 classes per subsequent task. The accuracy of each class in the first task is reported. From the perspective of representation learning, the low accuracy of some classes in joint training results from their semantic similarity. In the context of CIL, the forgetting of old classes occurs due to the introduction of similar new classes~\cite{AnatomySimilarity}. This explains why imbalanced forgetting occurs. For example, let's consider there are two classes, \texttt{\{male, dog\}} in the first task, and the model learns a new class, \texttt{female}. The \texttt{male} class tends to forget more easily due to representation interference.

\begin{figure*}[t]
 \centering
 \includegraphics[width=1.0\linewidth]{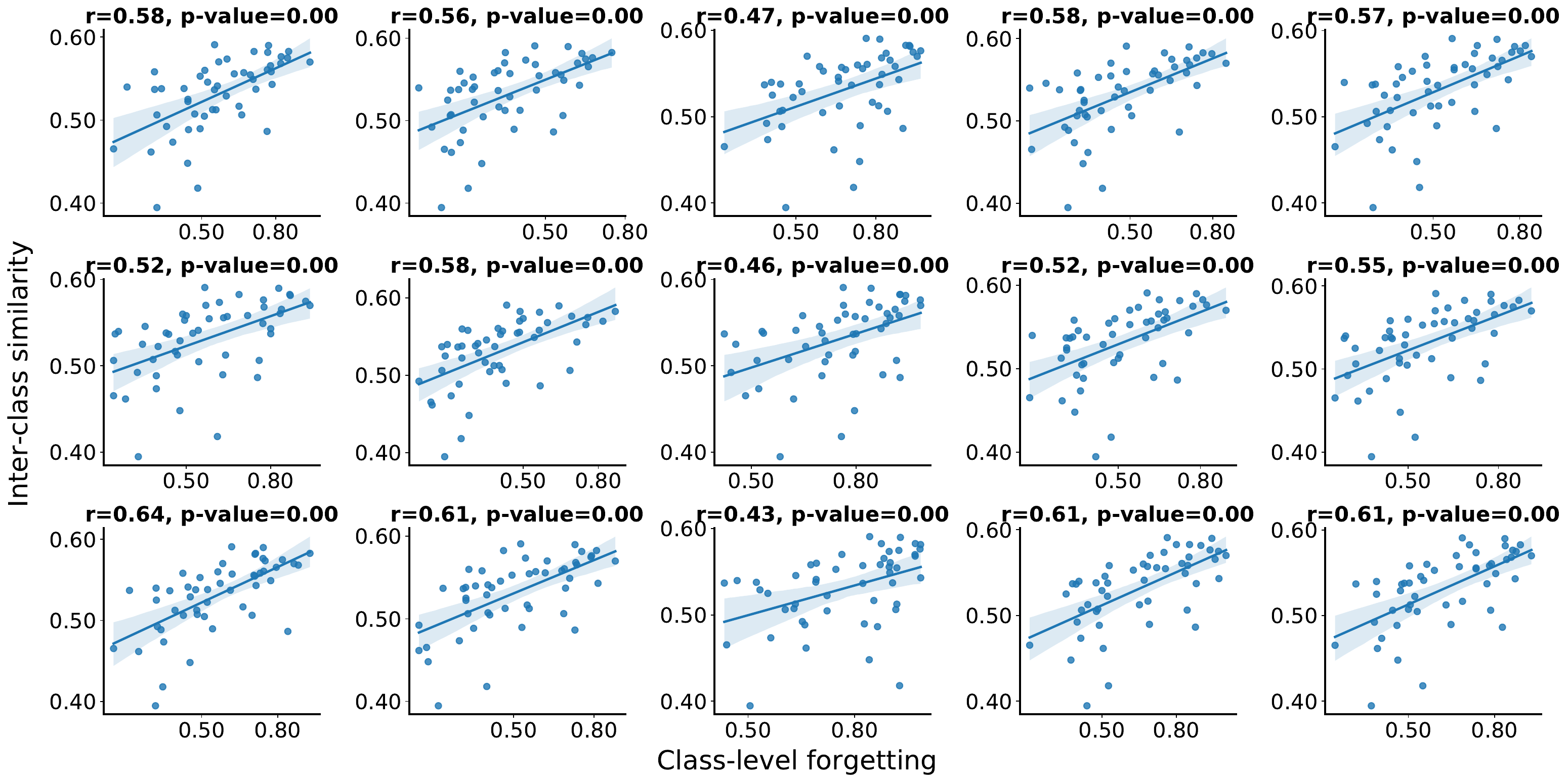}

 \caption{Illustration of the relative class forgetting and average similarity with latter classes for each old one. There is a positive correlation between the maximum similarity forgetting in different settings and methods. The first row gives the experiments that begin with 50 classes and 10 classes for each latter task, and the number of classes in the latter tasks in the second and third rows is 5 and 2, respectively.}
 \label{fig:method}
\end{figure*}

The above observations provide insight into the potential for improvement in average accuracy lying in the classes that are easy to forget. From a methodological perspective, the key is to minimize interference with the representations of those old vulnerable classes when new tasks arise. This raises two questions: 1) how to identify the vulnerable old classes and 2) how to mitigate conflicts between the representations of old and new classes. Regarding the first question, a positive relationship between inter-class similarity and class-level forgetting is established in conventional CIL settings. Therefore, it is possible to predict the forgetting level of old classes based on their similarity with the new classes. Building on this, a novel framework called \textit{CLass-Aware Disentanglement} (CLAD) is proposed to address the second question. As illustrated in Fig.~\ref{fig:overview} (b), CLAD consists of two phases: Forgetting Prediction (FP) and Representation Disentanglement (RD). In the FP phase, a subset of the old classes is identified as vulnerable ones, where conflicts are likely to occur with new classes. Empirical demonstrations and statistical analyses indicate a strong relationship between the conflict classes identified by FP and the degree of forgetting (see Fig.~\ref{fig:method}). During training for the new task, RD is introduced to constrain the similarity between the representations of samples in the new classes and the exemplars of their corresponding conflict classes. CLAD is formulated as a regularization term, which can be incorporated as a plugin for existing replay-based methods.

Extensive experiments on CIFAR-100~\cite{CIFAR100} and ImageNet~\cite{ImageNet} indicate that CLAD provides a consistent and impressive performance improvement over existing methods. Besides, comprehensive ablation studies are performed to show how the components in CLAD, including the conflict classes selection, regularization coefficient, and buffer size, influence its performance.

Our contributions are summarized as follows:
\begin{itemize}
 \item To the best of our knowledge, we are the first to reveal the imbalanced forgetting between the learned classes.
 \item Experiments and statistical analysis are conducted to demonstrate that imbalanced forgetting results from varying semantic similarity between inter-task classes.
 \item CLass-Aware Disentanglement (CLAD) is proposed to improve the accuracy of vulnerable old classes, which can be used as a plugin for replay-based CIL methods.
 \item Extensive experiments on several challenging benchmarks demonstrate that CLAD can provide consistent improvements over existing methods.
\end{itemize}

\section{Related Work}
\label{sec:related}
Typically, since only the data of the current task is available in each training phase, the main challenge of CIL is performance deterioration for old classes, {\it i.e.} catastrophic forgetting~\cite{catastrophic_forgetting}. An intuitive way is using a buffer to save some exemplars of each old class, and training the exemplars along with new data to mimic the i.i.d. joint training, named Experience Replay~\cite{ER, MER}. However, this strategy leads to a severe imbalance between current and old classes, and the bias towards new classes still exists~\cite{LUCIR, BiC, WA, GDumb}. 

Many approaches were proposed to further utilize the exemplars in recent years~\cite{LwF, LUCIR, ICARL, PODNet, GEM, AGEM, OGD, GPM, BiC, GDumb}, which can be split into three categories, knowledge distillation~\cite{LwF, LUCIR, ICARL, PODNet}, gradient projection~\cite{GEM, AGEM, OGD, GPM, FSDGPM}, and bias correction~\cite{LUCIR, BiC, WA, GDumb}, respectively. Knowledge distillation is a training strategy first proposed by~\cite{KD} to transfer the knowledge from the teacher model to the student model. LwF~\cite{LwF} uses this technology to preserve the knowledge of the old model (teacher model) during the new task training for the first time. Subsequent  methods~\cite{LUCIR, ICARL, PODNet} further involve the replay buffer and multiple distillation loss in CIL. Gradient projection methods try to keep the gradient of new tasks from interfering with old ones through projection, represented by GEM~\cite{GEM}, A-GEM~\cite{AGEM}, GPM~\cite{GPM}, FSDGPM~\cite{FSDGPM}, and OGD~\cite{OGD}. Inspired by the similarity between class imbalanced learning and CIL with exemplar replay~\cite{CILCIL}, BiC~\cite{BiC} and WA~\cite{WA} demonstrate that the bias occurs in the weights of classifier, and attempt to correct the bias by post-processing the weights. However, bias also exists in the backbone of the network. GDumb~\cite{GDumb} tackles this imbalance by constructing balanced data during training. 

Besides the end-to-end methods~\cite{LwF, LUCIR, PODNet, BiC}, several plugin methods emerging recently for CIL are based on the observation of the i.i.d. training process or results~\cite{CwD, CSCCT, AANet}. CwD~\cite{CwD} enforces the data representations to be more uniformly scattered at the first task, which mimics the representation extracted by the model trained with all classes (oracle model). CSCCT~\cite{CSCCT} proposes two regularization terms to cluster and distillate the class features, and encourage new classes to be situated optimally in the feature space. AANet~\cite{AANet} proposes to use a new branch for stable knowledge learning, which is effective but needs more memory.

\noindent\textbf{Discussion.} Most of the methods treat all the old classes equally and attempt to provide end-to-end methods to overcome the catastrophic forgetting problem. Differently, we try to mitigate the class-level representation interference in a class-aware way, which is inspired by our observation that the forgetting of different old classes is severely imbalanced. A similar effort has been made by LUCIR~\cite{LUCIR}, which adopts a margin ranking loss to encourage a large margin between the logits of old and new classes. However, it calculates the similarity for each sample, which will lead to inconsistent class-level similarity prediction. And it is not conducive to clustering together the representations of the same class. Our proposed Class-Aware Disentanglement (CLAD) measures the representation interference with cosine similarity and calculates the class similarity at the class level, which is more robust for CIL.

\section{Methodology}
\label{sec:method}

\subsection{Preliminary}
\label{sec:preliminary}

Formally, the sequentially trained model is denoted by $f(\cdot)=g(\phi (\cdot))$, which consists of a feature extractor $\phi(\cdot)$ followed by a classification layer $g(\cdot)$. $f(\cdot)$ is trained on sequential tasks $\mathcal{T}=[\mathcal{D}_1, \mathcal{D}_2, \ldots, \mathcal{D}_T]$ with non-overlapping classes. At the $t$-th training task, the model after training on the $(t-1)$-th task is denoted by $f_{t-1}$ and will be incrementally trained on the new dataset $\mathcal{D}=\mathcal{D}_{t}\cup \mathcal{B}$. The buffer $\mathcal{B}\subset [\mathcal{D}_1, \mathcal{D}_2, \ldots, \mathcal{D}_{t-1}]$ keeps only several exemplars of each old class. Under this setting, the standard cross-entropy loss of replay-based CIL methods is formulated as:
\begin{equation}
\label{equation:ce}
\mathcal{L}_{ce}=-\frac{1}{\mid \mathcal{D} \mid}\sum^{\mid \mathcal{D}\mid}_{k=1}y_k \log (\sigma(f(x_k))),
\end{equation}
where $(x_k,y_k)$ is an image and its label in $\mathcal{D}$. 

However, as the training data is highly imbalanced, $L_{ce}$ is not a good approximation of standard joint training classification loss at the $t$-th task.
Therefore, various additional constraints are proposed to help CIL methods better approximate the ideal loss during incremental training, such as knowledge distillation~\cite{LwF, ICARL, LUCIR}, re-sampling~\cite{GDumb, BiC, WA}, and gradient projection~\cite{GPM, AGEM, FSDGPM}. In general, denote these additional constraints by $L_{ad}$, the complete loss function $\mathcal{L}_{replay}$ of replay-based methods is formulated as:
\begin{equation}
 \mathcal{L}_{replay}=\mathcal{L}_{ce} + \lambda \mathcal{L}_{ad},
\end{equation}
where $\lambda$ stands for the adjustable weight of $\mathcal{L}_{ad}$.

\subsection{What Causes Imbalanced Forgetting?}
\label{sec:similarityforgetting}
In this section, we try to reveal the relationship between inter-class similarity and class-level forgetting. Intuitively, semantically similar classes are more likely to be misclassified from each other in standard model training with i.i.d. data. In the CIL setting with replay buffer, we can further infer that \textit{given a class the in old task, if there are classes very similar to it in the latter tasks, it is more likely to suffer performance degradation and vice versa}. 

To verify this hypothesis, three task sequences with different lengths are constructed using CIFAR-100~\cite{CIFAR100}. All three sequences begin with a base task containing 50 classes, and the number of classes in the subsequent tasks is 10, 5, and 2, respectively. Then 12 models are trained on the above three tasks with four representative CIL methods: naive replay with loss function defined in Eq.~\ref{equation:ce}, LUCIR~\cite{LUCIR}, BiC~\cite{BiC}, and iCaRL~\cite{ICARL}. The hyperparameters of all the experiments are consistent with the original papers, and the number of exemplars for each class is 20 as the common practice in~\cite{ICARL, LUCIR, AANet, PODNet}. The accuracy of each class after learning each task is recorded. Formally, we denote the accuracy of class $i$ after training on the \textit{base} task by $A_{base}^{i}$ and that after training on the \textit{all} tasks is denoted as $A_{all}^{i}$. As $A_{base}^{i}$ is different for each class $i$, the class-level forgetting $\delta_i$ is defined by normalized accuracy drop:
\begin{equation}
    \delta_i=\frac{A_{base}^i-A_{all}^i}{A_{base}^i}.
\end{equation}

To explore the relationship between the class-level forgetting $\delta_i$ and inter-class similarity, a reasonable approach is needed to measure the similarities between class $i$ and the latter 50 classes. A straightforward way is to train an \textbf{oracle} model on all the classes or even larger datasets, then use the features extracted by it to calculate inter-class cosine similarity~\cite{CwD}. Although this pipeline is standard and widely used in other fields like image retrieval~\cite{DeepIR} and multi-modal learning~\cite{multimodal}, two flaws limit its use in the CIL:
\begin{itemize}
    \item In the CIL setting, an oracle model is unavailable even after seeing all the classes.
    \item The representation space of the oracle model obtained by joint training is highly different from the changing one in incremental learning. 
\end{itemize}

To address the above limitations, we opt to use the model trained on the first 50 classes (denoted by $f_1$) instead of the oracle model for similarity calculation. The model $f_1$ is available after training of the first task, and the feature space aligns well with all the learned
50 classes naturally. Furthermore, the logits of the latter 50 classes obtained by $f_1$ are used as the similarity between old and new classes, which is essentially equivalent to the cosine similarity without normalization but much simpler in terms of calculation and implementation. Formally, the inter-class similarity level of each old class is defined as $S=Mean(f_1(C))$, where $C$ indicates a given new class. The inter-class similarity for old class $i$ and class C is denoted by $S_i=S[i]$. The functional equivalence between cosine similarity and logits similarity is shown in Tab.~\ref{tab:exp_metrics}.

With the above preparation, the relationship between inter-class similarity $S_i$ and class-level forgetting $\delta_i$ could be established. As shown in Fig.~\ref{fig:method}, there is an obvious positive correlation between the two variables under different settings and baselines. \textbf{Concretely, the Pearson correlation of them reached 0.6 with high confidence (p-value=0.00)}. This reveals that in the replay-based method, the representation of the class in the previous task might clash with the class representation in the new task similar to it, resulting in more pronounced forgetting. 
We would like to emphasize again that the purpose of this subsection is to \textit{establish} the relationship to find a method or a metric to predict where forgetting will happen.

\subsection{Class-Aware Disentanglement}
\label{sec:cad}
Motivated by the above observations and~\cite{AnatomySimilarity}, given a new class from the current task, encouraging its representation to keep distance from the old similar classes is beneficial to alleviate the forgetting of the corresponding old classes. To achieve this, two issues need to be addressed. First, how to predict classes that are most likely to be forgotten in the task sequence. Second, how to effectively separate the representations of the corresponding classes during the training process. In the subsequent content, the above two questions will be answered in turn. And the overview of the proposed CLAD is shown in Fig.~\ref{fig:overview} (b).

\noindent\textbf{Forgetting Prediction (FP).}
As stated above, we could predict which old classes in the first task are more likely to be forgotten with the model $f_1$. Extending this idea to the whole task sequence, the learned classes that are vulnerable to forgetting after the $(t-1)$-th task can be predicted by model $f_{t-1}$. Formally, for a new class $C$ in task $t$, the similarity between it and all the old classes is formulated as:

\begin{equation}
  \label{equation:logits_sim}
  S(C) = \frac{1}{\mid C \mid} \sum_{k=1}^{\mid C \mid}f_{t-1}(x_k),
\end{equation}
where $S(C)$ is the mean logit vector of class $C$ obtained by the model $f_{t-1}$. $|C|$ indicates the number of samples in class $C$ and $x_k$ is the $k$-th sample in class $C$. By sorting the $S(C)$, we can predict which old classes are most likely to forget when learning new class $C$.

\begin{table*}[htpb]
\small
  \centering
  \resizebox{\textwidth}{!}{
  \begin{tabular}{lcccccccc}
    \toprule
    
    \multirow{2}{*}{Method} & \multicolumn{3}{c}{CIFAR-100 ($B$=50)} & \multicolumn{3}{c}{ImageNet-100 ($B$=50)} &
    \multicolumn{2}{c}{ImageNet ($B$=100)} \\ 
    \cmidrule(l){2-4}\cmidrule(l){5-7}\cmidrule(l){8-9} & $S$=10 & 5 & 2 & 10 & 5 & 2 & 100 & 50 \\
    \midrule
    LwF~\cite{LwF} & 54.01 & 48.40 & 45.49 & 54.22 & 48.95 & 43.29 & 41.42 & 28.31 \\
    iCARL~\cite{ICARL} & 67.16 & 60.54 & 54.50 & 72.03 & 68.23 & 59.54 & 49.88 & 42.52 \\
    BiC~\cite{BiC} & 63.11 & 56.27 & 48.83 & 70.09 & 64.88 & 57.82 & 52.46 & 47.30 \\
    \midrule
    LUCIR~\cite{LUCIR} & 66.16 & 60.43 & 52.22 & 70.40 & 67.19 & 62.86 & 52.47 & 47.55 \\ 
    \quad + CLAD & 67.57$_{\bf +1.41}$ & 62.15$_{\bf +1.72}$ & 53.51$_{\bf +1.29}$ & 73.05$_{\bf +2.65}$ & 68.34$_{\bf +1.15}$ & 64.24$_{\bf +1.38}$ & 53.36$_{\bf +0.89}$ & 48.79$_{\bf +1.24}$ \\ 
    \midrule
    PODNet~\cite{PODNet} & 68.56 & 65.57 & 62.95 & 75.90 & 72.41 & 65.28 & 56.86 & 53.68  \\ 
    \quad + CLAD & 69.07$_{\bf +0.51}$ & 65.96$_{\bf +0.39}$ & 63.29$_{\bf +0.34}$ & 76.02$_{\bf +0.12}$ & 73.10$_{\bf +0.69}$ & 65.45$_{\bf +0.17}$ & 57.36$_{\bf +0.50}$ & 55.38$_{\bf +1.70}$  \\ 
    \midrule
    CwD~\cite{CwD} & 66.81 & 61.86 & 56.41 & 71.43 & 68.92 & 65.06 & 52.56 & 47.88 \\
    \quad + CLAD & 67.76$_{\bf +0.95}$ & 63.67$_{\bf +1.81}$ & 57.79$_{\bf +1.38}$ & 72.33$_{\bf +0.90}$ & 70.01$_{\bf +1.09}$ & 65.92$_{\bf +0.86}$ & 53.64$_{\bf +1.08}$ & 49.07$_{\bf +1.19}$ \\
    \bottomrule
  \end{tabular}}
  \setlength{\abovecaptionskip}{10pt}
  \caption{The improvement achieved by adding CLAD to the SOTAs~\cite{LUCIR,CwD,PODNet} and the comparison with three baselines~\cite{LwF,BiC,ICARL}. $B$ and $S$ denote the number of classes in the first task and the subsequent tasks. All the results are reproduced with the source code from~\cite{CwD}.
  }
  \label{tab:main_results}
\end{table*}

\noindent\textbf{Representation Disentanglement (RD).}
Equipped with FP, for a new class, we can locate which old classes are most likely to be forgotten. Naturally, forgetting can be mitigated by disentangling the learned representation of the new class and that of the corresponding old classes. 

Denoting the number of learned classes by $N$, a fixed proportion of the old classes are selected for new class $C$ as the conflict classes. This proportion $\mathcal{P}$ is defined as the conflict proportion. Given a sample $x$ from class $C$, the conflict classes are obtained by selecting the indexes of old classes that have the Top-$(\mathcal{P}*N)$ largest values in $S(C)$.

Now the conflict between the new classes and old classes can be mitigated by disentangling their representations. The new task is trained on joint data $\mathcal{D}=\mathcal{D}_{t}\cup \mathcal{B}$ with replay-based methods. In each iteration, the data batch contains both new and old classes. Based on this, a novel class-aware disentanglement regularization is proposed. Considering the representation distribution of the old classes is shifting during the new task training, the representation conflict is disentangled in both online and offline ways. In online disentanglement, new class representations are encouraged to be separated from the online representations of the conflict classes in the same batch. In offline disentanglement, the representations of class $C$ are encouraged to keep their distance from the old representations of all conflict classes in the buffer. Given a sample $x$ and its conflict samples $X_o$ in the same batch and $X_b$ in the buffer, the CLAD loss is formulated as:
\begin{equation}
  \mathcal{L}_{on}(x) = \frac{1}{\mid X_b \mid}\sum_{x_b \in X_b}(1+cos(\phi(x), \phi(x_b))),
\end{equation}
\begin{equation}
  \mathcal{L}_{off}(x) = \frac{1}{\mid X_o \mid}\sum_{x_o \in X_o}(1+cos(\phi(x), \phi_{t-1}(x_o))),
\end{equation}
where $cos(\cdot,\cdot)$ indicates the cosine similarity. Thus the objective of CLAD is:
\begin{equation}
  \mathcal{L}_{CLAD} = \frac{1}{\mid C \mid }\sum_{x \in C} (\mathcal{L}_\texttt{on}(x)+\mathcal{L}_\texttt{off}(x)).
\end{equation}
Accordingly, the overall loss function is written as:
\begin{equation}
  \mathcal{L} = \mathcal{L}_{replay} + \eta \mathcal{L}_{CLAD},
\end{equation}
where $\eta$ is the coefficient of CLAD loss.

\section{Experiments}
\label{sec:exp}

\subsection{Experimental Setup}
\label{sec:settings}
\noindent\textbf{Datasets and Protocols.} 
Three commonly used benchmarks~\cite{LUCIR} are selected to evaluate the proposed method. CIFAR-100~\cite{CIFAR100} consists of 600,000 images from 100 classes, and the image size is $32 \times 32$. ImageNet~\cite{ImageNet} contains about 1.2 million $224 \times 224$ RGB images from 1000 classes. ImageNet-100~\cite{ICARL} is a subset of ImageNet~\cite{ImageNet}, which is sampled as~\cite{LUCIR, AANet}. To be consistent with the protocols of the previous work~\cite{LUCIR, ICARL, AANet, Mnemonics, DistillingCausal}, all the classes of each dataset are shuffled with seed 1993 before splitting them into tasks. For CIFAR-100 and ImageNet-100, half classes are selected for the first task to mimic the pre-collected dataset in real-world~\cite{LUCIR}, then there are $S=10/5/2$ classes for each latter task. For ImageNet, 100 classes are selected for the first task, then the model learns 100 or 50 classes per task incrementally.

\noindent\textbf{Metrics.}
The average incremental accuracy ($A_t$)~\cite{PODNet, DistillingCausal, LUCIR} is used to evaluate the performance of the baselines and our results. Formally, denote the test accuracy of the model after the training of the $i$-th task as $A_i$, then the average incremental accuracy after the $t$-th task is defined as $A_t=\frac{1}{t}\sum_{i=1}^{t}A_i$.

\noindent\textbf{Implementation details.} Following the previous studies~\cite{CwD,DyER}, we adopt ResNet-18~\cite{ResNet} for all the experiments bellow. Notably, for CIFAR-100~\cite{CIFAR100} the kernel size of the first convolution layer is set to $3\times 3$, and the following maxpooling layer is removed for higher feature resolution~\cite{CwD}. And SGD is used as the optimizer. The learning rate is set to 0.1, the batch size is set to 128, the momentum is set to 0.9, and the weight decay is 5e-4. For CIFAR-100, all the methods are trained for 160 epochs for each task, and the learning rate is multiplied by 0.1 at the $80$-th and $120$-th epoch. For ImageNet and ImageNet100, the models are trained for 90 epochs for each task, and the learning rate is multiplied by 0.1 as the $30$-th and $60$-th epoch. Since we focus on the replay-based methods, the {\it Herding} strategy is used to select the exemplars for replay after training each task~\cite{ICARL}, and the number of exemplars per class is 20, which is consistent with~\cite{LUCIR, CwD}. The conflict proportion is set to 0.1 for all experiments empirically. The CLAD coefficient is set to 4 for LUCIR and CwD, while the value of it is 2 for PODNet. How to determine these values is detailed in the supplementary material.

\begin{figure*}[htbp]
  \centering
   \includegraphics[width=0.95\linewidth]{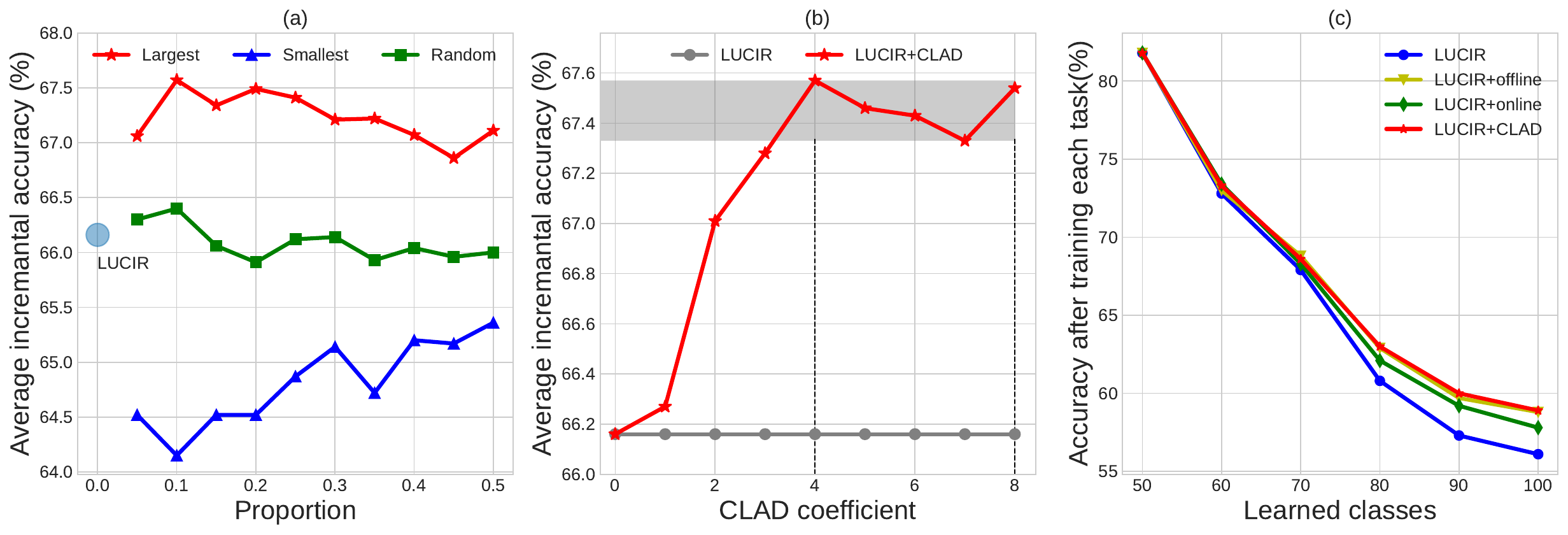}

   \caption{Ablation studies on the effectiveness of conflict prediction (a), the proportion of conflict classes (a), the impact of coefficient of CLAD loss (b), and components in conflict mitigation (c). The average incremental accuracy is reported for each experiment, which is averaged on three runs with different seeds.}
   \label{fig:ablation}
\end{figure*}

\subsection{Improvements over Baselines}
\label{sec:main_results}
We add our proposed CLAD to three strong CIL baselines: LUCIR~\cite{LUCIR}, CwD~\cite{CwD} and PODNet~\cite{PODNet}. Tab.~\ref{tab:main_results} shows the results on CIFAR-100, ImageNet-100, and ImageNet. Our method provides consistent improvement of average incremental accuracy by around $0.5\%$ and $2.5\%$ on various datasets and settings, \textit{e.g.}, on CIFAR-100, LUCIR with CLAD gains up to $1.72\%$ on accuracy while CwD with CLAD improves the baseline by $1.81\%$ at most. On ImageNet-100 when $C=10$, CLAD makes the LUCIR~\cite{LUCIR} even outperform the stronger baseline CwD~\cite{CwD}. CLAD also provides similar performance improvements on larger datasets ImageNet, indicating that our method adapts well to larger datasets. Besides, one may notice that the improvements on PODNet~\cite{PODNet} are limited compared with those on LUCIR and CwD. This phenomenon can be explained by the special design of PODNet, which conducts feature distillation even for middle layers to help preserve knowledge. However, our CLAD loss only disentangles the features in the final layer, which is consistent with the distillation loss in LUCIR and CwD. Although distillation does a good job of mitigating forgetting, the distillation with new samples will enhance the representation conflict in the middle layer in PODNet~\cite{forgettingless}.

\subsection{Ablation Study}
\label{sec:ablation}
In this subsection, extensive ablation studies are conducted to analyze the key components of CLAD. If not specified, all the experiments are based on LUCIR~\cite{LUCIR} under the protocol that split CIFAR-100~\cite{CIFAR100} into six tasks with 50 classes for the first task and 10 classes for the rest. The results are averaged over three runs.

\noindent\textbf{Effectiveness of conflict prediction.}
Although we are committed to mitigating representation conflict between similar old and new classes, it is doubtful that mitigating conflicts between arbitrary old and new classes also help CIL. To dispel this concern, apart from selecting the Top-$\mathcal{P}N$ largest values in $S(C)$ as old classes, we select the Top-$\mathcal{P}N$ smallest values in $S(C)$ and randomly select $\mathcal{P}N$ old classes for comparison, named as \textit{Smallest} and \textit{Random}. As shown in Fig.~\ref{fig:ablation} (a), the latter two strategies are harmful no matter what proportion of the old classes is chosen. Because the CLAD loss constrains the feature distribution of the new class to some extent. But the compromise of new classes will not benefit the performance of the old class when there is no representation conflict between them, which happens in \textit{Smallest} and \textit{Random} strategies.

\noindent\textbf{Proportion of conflict classes.}
In this part, we further investigate the impact of different proportions of chosen conflict classes for each new class. In Fig.~\ref{fig:ablation} (a), it is shown that the improvement gained by CLAD is relatively small when the proportion is extremely large or small, while CLAD achieves the greatest performance gain with a proportion set to 0.1. This observation is intuitive because too few conflict classes are not sufficient for mitigation and there can be prediction errors. Too many conflicting classes will hurt the performance of the new class more because some old classes do not conflict with the new class.

\noindent\textbf{Components in conflict mitigation.}
The proposed CLAD loss has two components as stated above, now we ablate each component of CLAD in Fig.~\ref{fig:ablation} (c). Each of the components can improve the average incremental accuracy of the baseline, and the combination of both can further help improve the performance.

\begin{table}[htbp]
 \centering
 \resizebox{\linewidth}{!}{
 \begin{tabular}{ccccccc}
 \toprule
 \multirow{2}{*}{$R$} & \multicolumn{3}{c}{$S$=10} & \multicolumn{3}{c}{$S$=5} \\ 
    \cmidrule(l){2-4}\cmidrule(l){5-7} & LUCIR & $w/$ CLAD & $\uparrow$ & LUCIR & $w/$ CLAD & $\uparrow$ \\
 \midrule
 5 & 55.65& 58.22 & \textbf{2.57} & 52.46 & 56.09 & \textbf{3.63}\\
 10 & 63.28 & 65.49 & 2.21 & 56.80 & 59.33 & 2.53\\
 20 & 66.16 & 67.57 & 1.41 & 60.43 & 62.15 & 1.72\\
 30 & 67.62 & 68.57 & 0.95 & 62.96 & 63.87 & 0.91\\
 40 & 68.58 & 69.07 & 0.49 & 63.51 & 64.32 & 0.81\\
 \bottomrule
 \end{tabular}}
 \setlength{\abovecaptionskip}{10pt}
 \caption{Ablation study on the number of exemplars. The number of exemplars per class is denoted by $R$.}
 \label{tab:ablation-buffer}
\end{table}

\noindent\textbf{Improvement with different exemplar numbers.}
We attempt to verify the effectiveness of CLAD by testing it with varying numbers of exemplars per class. The corresponding results are listed in Tab.~\ref{tab:ablation-buffer}. Notably, our approach produces increasingly significant performance gains as the number of exemplars decreases. This phenomenon suggests that a reduced number of exemplars per class exacerbates the imbalance of forgetting between old classes, making our approach particularly effective in tackling this challenging scenario.

\noindent\textbf{Measurements of conflict prediction.}
Various similarity measurements for conflict prediction are available for CLAD. But the logits-based similarity is sufficient for our proposed CLAD. To be more convincing, we compare the performance differences using different similarity measurements in Tab.~\ref{tab:exp_metrics}. It shows that there are no obvious differences in performance between the two measurements, and our approach is more concise and efficient. Furthermore, we experimentally prove that FP using the oracle model is ineffective, which supports our aforementioned analysis.
\begin{table}[htbp]

 \centering
 \resizebox{\linewidth}{!}{
 \begin{tabular}{lcccc}
 \toprule
 Similarity & LUCIR & \textit{oracle} & logits & cosine \\
 \midrule
 $A_{0.05}(\%)$ & 66.16 & 65.84$_{\bf{-0.32}}$ & 67.06$_{\bf +0.90}$ & 67.22$_{\bf+1.06}$ \\
 $A_{0.10}(\%)$ & 66.16 & 65.90$_{\bf{-0.26}}$ & 67.57$_{\bf +1.38}$ & 67.45$_{\bf +1.29}$ \\
 \bottomrule
 \end{tabular}
 }
 \setlength{\abovecaptionskip}{10pt}
 \caption{Ablation study on the different measurements for forgetting prediction. $A_{0.05}$ and $A_{0.10}$ denote the average incremental accuracy with conflict proportions of 0.05 and 0.10. \textit{logits} is the adopted measurement and \textit{cosine} is provided as an alternative. The result using an oracle model is also given as \textit{oracle}.}
 \label{tab:exp_metrics}
\end{table}

\noindent\textbf{Impact of coefficient of CLAD loss.}
We demonstrate the improvement with different coefficients of CLAD loss varying from 1.0 to 8.0 in Fig.~\ref{fig:ablation} (b). Interestingly, our method is not sensitive to this coefficient, especially when it is greater than 4.0. This phenomenon indicates that when the current class is dissimilar enough to the old classes in the buffer, a larger coefficient will not have a more significant effect. These results also reflect the robustness of CLAD.

\section{Conclusion}
\label{sec:conclusion}
We analyze catastrophic forgetting by revealing imbalanced forgetting in Class Incremental Learning (CIL). Extensive empirical studies and analyses are conducted to establish the connection between imbalanced forgetting and inter-class similarity. Based on this, a forgetting prediction method and a regularization term named CLAD are designed to disentangle the representation interference of similar old and new classes. The effectiveness of CLAD in improving existing methods is demonstrated across multiple experimental settings. Additionally, comprehensive ablation studies are conducted to verify the rationality of our design. This work provides a novel perspective of imbalanced forgetting in CIL, which might stimulate future research in this field. 

\noindent\textbf{Limitation.} There are also limitations to our CLAD that are worth further exploration. For example, other kinds of losses and old class selection methods may need to be explored. Numerous exemplar-free methods for CIL are not covered in this research. We plan to include them in our future work. 

\section{Acknowledgements}
This research was supported by the National Natural Science Foundation of China under Grants 62376267, 61976208, 62071466, 62076242, 62222302 and the InnoHK project.

\bibliography{aaai24}

\end{document}